\documentclass[11pt]{article}
\usepackage[margin=1in]{geometry}
\usepackage{xcolor,booktabs,tabularx,amsmath,setspace}
\usepackage{graphicx}
\usepackage{enumitem}
\usepackage{caption}
\usepackage{listings}
\usepackage[most]{tcolorbox}
\usepackage{float}   %
\definecolor{raiblue}{HTML}{1B5FAA}\definecolor{raipale}{HTML}{F3F7FC}\definecolor{raiedge}{HTML}{A9C4E6}
\newcommand{\promptvar}[1]{\textcolor{raiblue}{\textbf{\{#1\}}}}
\newtcblisting{promptbox}[1]{enhanced, breakable, listing only, arc=2mm, boxrule=0.5pt, colframe=raiedge, colback=raipale,
  colbacktitle=raiblue, coltitle=white, fonttitle=\sffamily\small\bfseries, title={#1}, left=6pt, right=6pt, top=4pt, bottom=4pt,
  before skip=6pt, after skip=10pt, listing options={basicstyle=\scriptsize\ttfamily, breaklines=true, breakindent=0pt,
  columns=fullflexible, keepspaces=true, breakatwhitespace=true, upquote=true, escapeinside={(*@}{@*)}}}
\newtcolorbox{promptfig}[1]{colback=raipale, colframe=raiblue, coltitle=white, title={#1}, width=\linewidth, fontupper=\footnotesize}
\usepackage[numbers,round,sort&compress]{natbib}  %
\usepackage{hyperref}
\hypersetup{colorlinks=true, linkcolor=raiblue, citecolor=raiblue, urlcolor=raiblue}

\newcommand{\new}[1]{#1}   %
\newcounter{appx}\renewcommand{\theappx}{S\arabic{appx}}
\newcommand{\appsection}[2]{\refstepcounter{appx}\label{#2}\subsection*{Appendix \theappx: #1}}

\newcolumntype{Y}{>{\centering\arraybackslash}X}
\definecolor{dgreen}{RGB}{56,142,60}\definecolor{dred}{RGB}{140,30,30}
\newcommand{\dup}[1]{\,{\tiny\textcolor{dgreen}{$\uparrow$#1}}}
\newcommand{\ddown}[1]{\,{\tiny\textcolor{dred}{$\downarrow$#1}}}

\begin{document}

\begin{center}
{\Large\bfseries CTLink: A Text-Prompted Segmentation Model for Linking Radiology Report Findings to 3D Chest CT Scans}\\[10pt]
{\small \mbox{Mohammed Baharoon\textsuperscript{1,*}}, \mbox{Sung Eun Kim\textsuperscript{1}}, \mbox{Mahmoud Hussain Alabbad\textsuperscript{2}}, \mbox{Lucas Bijnens\textsuperscript{3}}, \mbox{Ali Piracha\textsuperscript{4}}, \mbox{Reem Alwasiah\textsuperscript{5}}, \mbox{Talal Yahyan\textsuperscript{5}}, \mbox{Abdulrahman Alamoudi\textsuperscript{5}}, \mbox{Mashael Alshehri\textsuperscript{5}}, \mbox{Rayan Alsager\textsuperscript{5}}, \mbox{Hannah Hancock\textsuperscript{6}}, \mbox{Anna M. Hunkem\"{o}ller\textsuperscript{7,8}}, \mbox{Peter Youssef\textsuperscript{9}}, \mbox{Mona A. Alhammad\textsuperscript{10}}, \mbox{Afnan Aljumayan\textsuperscript{10}}, \mbox{Brecht Van Berkel\textsuperscript{11}}, \mbox{Eric Bijnens\textsuperscript{11}}, \mbox{Saud Alhussain\textsuperscript{12}}, \mbox{Maarten Steyvers\textsuperscript{13}}, \mbox{Rolf Symons\textsuperscript{14}}, \mbox{Bryan Eelen\textsuperscript{3}}, \mbox{Ayed Alaklubi\textsuperscript{15}}, \mbox{Maha Buraykan Alshammari\textsuperscript{12}}, \mbox{Amirah Fahad Alshammeri\textsuperscript{16}}, \mbox{Tomas Gilvydis\textsuperscript{4}}, \mbox{Ibrahim AlTurki\textsuperscript{17}}, \mbox{Monerah Almohideb\textsuperscript{17}}, \mbox{Ali Alburkani\textsuperscript{5}}, \mbox{Shahad Alhomaiani\textsuperscript{18}}, \mbox{Fatima Jabali\textsuperscript{19}}, \mbox{Abdulrhman Aljouie\textsuperscript{20}}, \mbox{Daesung Kim\textsuperscript{21}}, \mbox{Junha Im\textsuperscript{22}}, \mbox{Michael Moritz\textsuperscript{4,23}}, \mbox{Hassan R. Alomaish\textsuperscript{5}}, \mbox{Pranav Rajpurkar\textsuperscript{1}}}\\[8pt]
\begin{minipage}{\linewidth}\footnotesize
\textsuperscript{1}Department of Biomedical Informatics, Harvard Medical School, Boston, MA, USA; \textsuperscript{2}King Fahad Hospital, Al-Ahsa Health Cluster, Al Hofuf, Saudi Arabia; \textsuperscript{3}KU Leuven, Leuven, Belgium; \textsuperscript{4}Saint Louis University, St Louis, MO, USA; \textsuperscript{5}Medical Imaging Department, King Abdulaziz Medical City, Ministry of National Guard Health Affairs, Riyadh, Saudi Arabia; \textsuperscript{6}Naval Medical Center Portsmouth, Portsmouth, VA, USA; \textsuperscript{7}Institute of Diagnostic and Interventional Radiology, Hannover Medical School, Hannover, Germany; \textsuperscript{8}Biomedical Research in Endstage and Obstructive Lung Disease Hannover (BREATH), German Center for Lung Research (DZL), Hannover, Germany; \textsuperscript{9}Justus Liebig University Giessen, Giessen, Germany; \textsuperscript{10}Ras Tanura General Hospital, Ministry of Health, Ras Tanura, Saudi Arabia; \textsuperscript{11}Jessa Hospital, Hasselt, Belgium; \textsuperscript{12}King Fahad Specialist Hospital, Buraydah, Saudi Arabia; \textsuperscript{13}University Hospitals Leuven, Leuven, Belgium; \textsuperscript{14}Imelda Hospital, Bonheiden, Belgium; \textsuperscript{15}Prince Sultan Military Medical City, Riyadh, Saudi Arabia; \textsuperscript{16}University of Hail, Hail, Saudi Arabia; \textsuperscript{17}King Saud University Medical City, Riyadh, Saudi Arabia; \textsuperscript{18}King Abdulaziz Specialist Hospital, Taif Health Cluster, Taif, Saudi Arabia; \textsuperscript{19}King Abdulaziz University Hospital, Jeddah, Saudi Arabia; \textsuperscript{20}Research and Innovation Institute, Ministry of Defense Health Services, Riyadh, Saudi Arabia; \textsuperscript{21}Yonsei University, Seoul, South Korea; \textsuperscript{22}Yonsei University College of Medicine, Seoul, South Korea; \textsuperscript{23}SSM Health, St Louis, MO, USA.
\end{minipage}
\end{center}
{\let\thefootnote\relax\footnotetext{\textsuperscript{*}Corresponding author: Mohammed Baharoon (\href{mailto:mohammed_baharoon@hms.harvard.edu}{mohammed\_baharoon@hms.harvard.edu}).}}

\begin{abstract}
Radiology reports describe findings in text only, without visually marking them on the scan, so readers cannot
directly see which abnormality each reported finding refers to.
We introduce \textbf{CTLink}, a text-prompted 3D segmentation model that segments findings described in a
chest CT report. CTLink was trained on 4123 noncontrast
chest CT scans with 9547 segmented report findings, including ReXGroundingCTv2, a newly annotated dataset of 1134 scans
and 1895 findings with every instance of each finding segmented. On the 200 ReXGroundingCT validation scans, CTLink achieved a hit
rate of 0.816, an instance F1 score of 0.456, and a mean Dice score of 0.400, compared with 0.533, 0.175, and 0.234
for VoxTell, a publicly available text-prompted segmentation model ($P < .001$ for all). On the test set of the MICCAI
2026 ReXGrounding challenge, an ensemble of seven CTLink models achieved the highest hit rate (0.782) and instance F1
score (0.386) of all 48 participating teams. On 50 test scans segmented twice by independent radiologists, CTLink matched or exceeded inter-rater agreement. Diffuse
non-focal findings remained the main limitation, partly because their extent is outlined
inconsistently even between radiologists. The CTLink model weights are available at
\url{https://huggingface.co/rajpurkarlab/CTLink}, and the dataset at
\url{https://huggingface.co/datasets/rajpurkarlab/ReXGroundingCT}.
\end{abstract}

\section*{Introduction}
Radiologists record the findings of an imaging examination in a radiology report, which describes each finding in
text only, without marking it on the scan. Referring clinicians who want to see a described finding must therefore locate it on the scan themselves, which takes time and can be ambiguous when a scan contains several similar abnormalities.
When findings are visually marked on the scan, referring clinicians plan management faster and more confidently
\cite{iyer2010added}.
A model that segments the finding described by each report sentence would automatically highlight that finding on the scan,
linking every sentence to the abnormality it describes. Such a model would spare referring clinicians the
search for each described finding, make it easier to compare findings with those on prior examinations, and make
AI-generated reports more verifiable
\cite{rajpurkar2023current,bannur2024maira,baharoon2026rexgroundingct}.

For three-dimensional scans such as chest CT, the task of segmenting the findings described by report sentences remains difficult \cite{baharoon2025state}. A report sentence
specifies a finding by its appearance, size, and location, often among several similar abnormalities in the same
scan. Most segmentation models, however, are trained to segment a fixed set of organs or lesion types
\cite{isensee2024nnu,wasserthal2023totalsegmentator}, accepting only a category name, such as ``nodule'' or
``tumor,'' as the prompt \cite{zhao2025large,zhao2025foundation,liu2023clip,du2024segvol}. Such models perform poorly on free-text report sentences
\cite{baharoon2025state}, because these sentences differ from the short category names
the models were trained on. Linking phrases to the image regions they describe, a task known as visual grounding
\cite{plummer2015flickr30k,hu2016segmentation}, has been studied in radiology mainly on chest radiographs, with bounding boxes rather than segmentations \cite{boecking2024ms,wu2021chest,muller2024chex,de2025padchest,bannur2024maira}. %
More recent work on CT has linked report text to anatomical regions, such as organs or lung lobes \cite{zhang2025development,chen2025large}, or has localized findings with bounding boxes \cite{ichinose2023visual}, rather than segmenting the described findings. Progress has long been limited by the lack of a public dataset that pairs report sentences with segmentations of the described findings. ReXGroundingCT now provides such data, linking findings from chest CT reports \cite{hamamci2026generalist} to 3D segmentation masks
\cite{baharoon2026rexgroundingct}. %
Models trained with ReXGroundingCT, such as VoxTell \cite{rokuss2026voxtell} and DAGG
\cite{uhm2026decouple},
have substantially advanced performance on this task but have not reached the level of agreement between radiologists
\cite{baharoon2026rexgroundingct}.

We introduce CTLink, a text-prompted segmentation model for linking radiology report findings to 3D chest CT scans. To train it, we create ReXGroundingCTv2, which adds 1134 chest CT scans with every instance of each finding segmented to ReXGroundingCT.
On the ReXGroundingCT validation set, CTLink localizes 82\% of findings, compared with 53\% for VoxTell. %
On 50 test scans segmented twice by independent radiologists
\cite{baharoon2026rexgroundingct}, CTLink matches or exceeds inter-rater agreement.
On the test set of the ReXGrounding challenge at MICCAI 2026, an ensemble of CTLink
models localizes more findings (78\%) and achieves a higher instance F1 score (0.39) than the entries of all 48
participating teams \cite{baharoon2026grounding}.
We publicly release the CTLink model weights and ReXGroundingCTv2 to further support research in this direction.

\section*{Materials and Methods}

\subsection*{Datasets and Annotations}
This retrospective study used de-identified noncontrast chest CT scans and radiology reports from CT-RATE
\cite{hamamci2026generalist}, a public dataset acquired during routine care at Istanbul Medipol University Mega
Hospital, Turkey, and released under its own ethics approval.
Additional institutional review board approval was not required.
Table~\ref{tab:data} summarizes the datasets used for training and evaluation. The training set of the original ReXGroundingCT (v1 training set) contained 2992 scans and
7687 findings, with at most three instances annotated per finding \cite{baharoon2026rexgroundingct}. We used the
corrected v1 masks released by Team 4 of the ReXGrounding challenge (D.K., J.I.), who made 146 changes to 117 scans
(97 masks redrawn, 35 findings removed, and 14 other edits, such as sentence fixes and side corrections), leaving 2989
scans and 7652 findings \cite{baharoon2026grounding}. As part of ReXGroundingCTv2, we annotated 1134 additional
training scans with 1895 findings (v2 training set). The 200 ReXGrounding
challenge validation scans (381 findings) were used for the ablation, and the 300 ReXGrounding challenge test scans (582
findings) were used for the comparison with the challenge entries. Eight validation scans and 15 test scans came from patients who also had a
different examination among the scans used for training.

\begin{table}[!htbp]
\small
\caption{\textbf{Datasets.} All scans are noncontrast chest CT scans from CT-RATE. ReXGroundingCTv2 adds the v2 training scans to ReXGroundingCT. *With the corrected masks released by Team 4 of the ReXGrounding challenge \cite{baharoon2026grounding}. The original v1 training set has 2992 scans and 7687 findings.}
\begin{tabularx}{\linewidth}{@{}Xcccccc@{}}
\toprule
Dataset & Scans & Findings & Focal & Non-focal & Annotation & Use \\
\midrule
v1 training set* & 2989 & 7652 & 6055 & 1597 & Up to three instances & Training \\
v2 training set & 1134 & 1895 & 1704 & 191 & Every instance & Training \\
Validation set & 200 & 381 & 329 & 52 & Every instance & Ablation \\
Test set & 300 & 582 & 509 & 73 & Every instance & Challenge comparison \\
\bottomrule
\end{tabularx}
\label{tab:data}
\end{table}

Finding sentences came from the CT-RATE reports, which had been translated from Turkish into English. For
ReXGroundingCT, a large language model rewrote each report into standard terminology and split it into individual
findings \cite{baharoon2026rexgroundingct}. Only findings of the lungs, airways, and pleura were
kept, so findings of other structures, such as the mediastinum or bones, were not annotated.
Each finding sentence had one
reference mask, drawn on a cloud-based annotation platform with brush, polygon, and region-growing tools, in which
each instance of the finding (for example, each nodule) was labeled separately. %
The v1 training masks included at most three instances of each finding, whereas the v2 masks, just like the validation and test masks,
included every visible instance. The annotation process is described in more detail in Appendix~\ref{app:annot}. Findings that could not be localized,
were too diffuse to outline, or described normal anatomy were not annotated.

\subsection*{Model}
\begin{figure}[!htbp]\centering
\includegraphics[width=\linewidth]{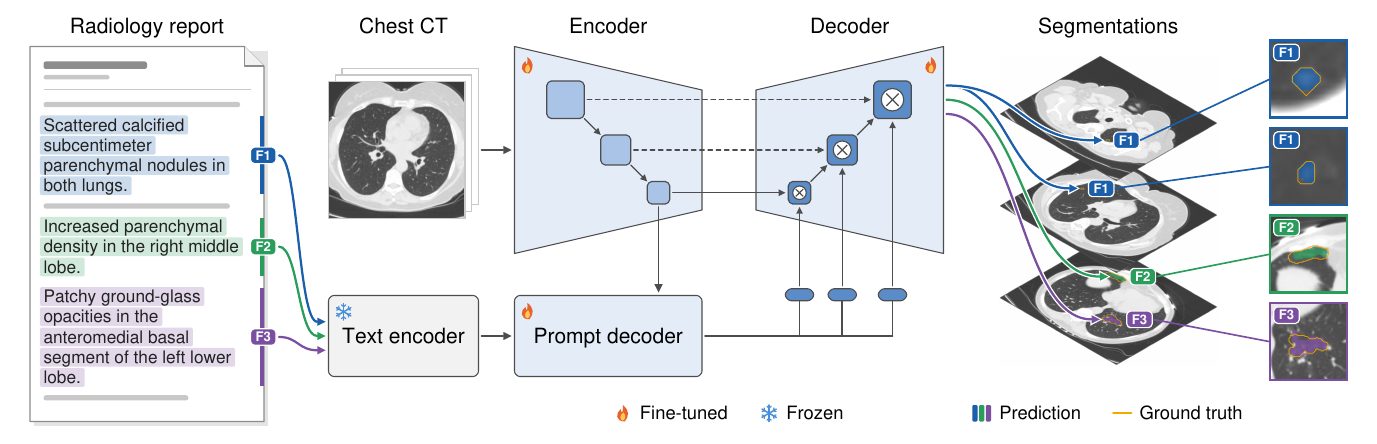}
\caption{Architecture of CTLink, which follows that of VoxTell \cite{rokuss2026voxtell}. Each finding sentence of the report
is embedded by a frozen text encoder, and the chest CT scan is encoded by a U-Net encoder. A prompt decoder combines the
sentence with the image features, and its output is fused with the decoder features at each resolution ($\otimes$) to
segment the finding.}
\label{fig:overview}
\end{figure}

CTLink used the same architecture as VoxTell v1.1 \cite{rokuss2026voxtell}: a residual-encoder 3D U-Net \cite{isensee2024nnu} image
encoder and a transformer prompt decoder. In the prompt decoder, the sentence embedding, computed by a frozen Qwen3-Embedding-4B
model \cite{zhang2025qwen3}, attended to the image features, and the resulting text features were fused with the decoder features at every stage. Each finding sentence was one
prompt, and the output was a binary mask in the native CT voxel grid. Figure~\ref{fig:overview} shows the architecture.

\subsection*{Training}
All models were fine-tuned from the VoxTell v1.1 weights. This version was trained on 190 public datasets and drew only 5\% of its training samples from instance-focused datasets such as ReXGroundingCT \cite{rokuss2026voxtell}, so report sentences were a small part of its training. Each training sample was one scan--finding pair,
with a random 3D crop of $192^3$ voxels that was centered on the finding in 50\% of samples. Training used
a combined Dice and binary cross-entropy loss with deep supervision and the AdamW optimizer (learning rate $1\times10^{-5}$, 500-iteration
linear warm-up, cosine decay) for 50\,000 iterations with a batch size of two (100\,000 training samples). For v1 findings with three annotated
instances, background voxels were down-weighted (weight, 0.3), because further instances may be unlabeled.

The ablation (Table~\ref{tab:ablation}) started from the published VoxTell model without fine-tuning and added one
change per row. Oversampling drew the v2 scans three times as often as the v1 scans, larger
input crops increased the crop size from $192^3$ to $192\times256\times256$ voxels, and longer training increased
the number of iterations from 50\,000 to 75\,000 and the effective batch size from two to four (300\,000 instead of
100\,000 training samples).
No ablation model was trained on validation scans.

\begin{table}[!htbp]
\footnotesize
\caption{\textbf{Cumulative Ablation on the Validation Set.} Data are values for the ReXGrounding challenge validation set. Rows are cumulative; arrows show the change from the row above (\textcolor{dgreen}{green}, increase; \textcolor{dred}{red}, decrease). The first row is the published VoxTell model without fine-tuning. Oversampling of v2 draws the v2 scans three times as often as the v1 scans, larger input crops measure $192\times256\times256$ instead of $192^3$ voxels, and longer training uses 75\,000 instead of 50\,000 iterations and an effective batch size of four instead of two. Hit rate is the proportion of findings with a Dice score of at least 0.1, Dice is the mean per-scan Dice score, and instance F1 matches connected components of the prediction one to one with the annotated reference instances. The v1 scans are the original ReXGroundingCT training scans, and the v2 scans are the training scans annotated exhaustively for this study (added in ReXGroundingCTv2). *With the corrected training masks.}
\setlength{\tabcolsep}{3pt}
\begin{tabularx}{\linewidth}{@{}>{\raggedright\arraybackslash}p{0.34\linewidth}YYY@{}}
\toprule
Model & Instance F1 & Dice & Hit Rate \\
\midrule
VoxTell \cite{rokuss2026voxtell} & 0.175 & 0.234 & 0.533 \\
+ v1 scans* & 0.298\dup{0.123} & 0.353\dup{0.119} & 0.761\dup{0.228} \\
+ v2 scans & 0.316\dup{0.018} & 0.388\dup{0.035} & 0.793\dup{0.032} \\
+ Oversampling of v2 & 0.310\ddown{0.006} & 0.393\dup{0.005} & 0.803\dup{0.010} \\
+ Larger input crops & 0.328\dup{0.018} & 0.396\dup{0.003} & 0.801\ddown{0.002} \\
+ Longer training & 0.373\dup{0.045} & 0.403\dup{0.007} & 0.824\dup{0.023} \\
+ Post-processing (= CTLink) & 0.456\dup{0.083} & 0.400\ddown{0.003} & 0.816\ddown{0.008} \\
\bottomrule
\end{tabularx}
\label{tab:ablation}
\end{table}

For the comparison with the ReXGrounding challenge entries on the 300 challenge test scans, CTLink was retrained with the same
recipe on the training scans plus the 200 validation scans (sampled as often as v2 scans), with a batch size of four.
The final checkpoint was used, without any selection on test data. Models were trained on NVIDIA L40S (48 GB) or H100 (80 GB) GPUs. The CTLink model for the ReXGrounding challenge comparison was trained on one H100 GPU.

\subsection*{Post-processing}
We applied four post-processing steps. First, predictions were restricted to the lobes named in the sentence, using the lung lobe masks provided with CT-RATE (generated with
TotalSegmentator \cite{wasserthal2023totalsegmentator}) dilated by 32 voxels. Second, non-focal findings other than emphysema used a probability threshold of 0.9 instead of
0.5, because the model over-segmented these diffuse findings relative to the reference masks. Third, emphysema
predictions were kept only in lung with attenuation below $-910$ HU, plus a three-voxel margin, because emphysema appears
on CT as lung with abnormally low attenuation
\cite{muller1988objective}. Finally, connected components smaller than 5\% of the largest component of the finding were
removed. Post-processing settings were selected on the validation set and the public half of the ReXGrounding challenge
test set (150 scans).

\subsection*{CTLink Ensemble}
We also trained an ensemble of seven CTLink models as our entry for the ReXGrounding challenge comparison
(Table~\ref{tab:challenge}). The ensemble averaged the probability maps of its members, which were selected on the public half of the ReXGrounding challenge test set (150 scans), as was possible for teams
during the challenge's public leaderboard phase. The members were five trainings of the CTLink recipe with different random seeds and two variants, one trained with non-focal findings sampled three times as often and one fine-tuned for 10\,000 iterations without the non-focal findings of v1 scans. Post-processing was the same as for CTLink, except that, because averaging lowers the probabilities of the members, the mean probability map was thresholded at 0.35 instead of 0.5 (0.8 instead of 0.9 for non-focal findings other than emphysema), and predicted components smaller than 50 voxels were removed.

\begin{table}[!htbp]
\small
\caption{\textbf{Comparison with the ReXGrounding Challenge Entries on the Test Set.} Results for the other teams come from the ReXGrounding challenge paper \cite{baharoon2026grounding}. CTLink is one network with the post-processing described in Materials and Methods (lobe restriction, non-focal threshold, emphysema gating, and removal of small components), and the CTLink ensemble averages seven CTLink models, with post-processing. Instance and hit precision and recall are in Table~\ref{tab:supp_test}.}
\setlength{\tabcolsep}{3.5pt}
\begin{tabularx}{\linewidth}{@{}>{\raggedright\arraybackslash}p{0.46\linewidth}YYY@{}}
\toprule
Entry & Instance F1 & Mean Dice & Hit Rate \\
\midrule
CTLink without post-processing & 0.264 & 0.369 & 0.773 \\
CTLink & 0.378 & 0.364 & 0.766 \\
CTLink ensemble & \textbf{0.386} & 0.373 & \textbf{0.782} \\
\midrule
Team 1 (winner) & 0.341 & \textbf{0.375} & 0.768 \\
Team 2 & 0.343 & 0.365 & 0.763 \\
Team 3 & 0.334 & 0.362 & 0.765 \\
Team 4 & 0.312 & 0.361 & 0.777 \\
Team 5 & 0.190 & 0.353 & 0.746 \\
Team 6 & 0.243 & 0.352 & 0.765 \\
Team 7 & 0.261 & 0.343 & 0.753 \\
Team 8 & 0.246 & 0.339 & 0.742 \\
Team 9 & 0.248 & 0.333 & 0.716 \\
Team 10 & 0.299 & 0.333 & 0.689 \\
\bottomrule
\end{tabularx}
\label{tab:challenge}
\end{table}

\subsection*{Evaluation Metrics}
All metrics were computed with the official ReXGrounding challenge evaluator \cite{baharoon2026grounding}. Hit rate is the proportion of
findings with a Dice score of at least 0.1 between the predicted and reference masks. The instance F1 score measures whether the separate instances of a
finding are found, by matching connected components of the prediction one to one with the annotated reference instances (match at Dice
score $\geq$ 0.2).

\subsection*{Comparison with Radiologists}
In the inter-rater study of ReXGroundingCT \cite{baharoon2026rexgroundingct}, the findings of 50 cases from the test set
were segmented a second time, independently, by other board-certified radiologists. On the 96 findings that the second
segmentation also localized, we compared the agreement of CTLink with each segmentation with the agreement between the two segmentations
(Appendix~\ref{app:f1r2}).

\subsection*{Robustness Analysis}
To test the robustness of CTLink to phrasing, we evaluated it on edited versions of the 582 test finding sentences. A large
language model (GPT-6.1-sol; OpenAI) edited each sentence in five ways: (1) replacing terms, such as finding names or
locations, with synonyms that radiologists typically use in chest CT reports, (2) replacing terms with standard radiology abbreviations, (3) freely rewording the whole
sentence, (4) for findings in both lungs, splitting the sentence into a left-lung and a right-lung sentence, and (5) for
findings in two or more lobes of one lung, splitting the sentence into one sentence per lobe. Details of the process are given in Appendix~\ref{app:rob}.

\subsection*{Error Analysis}
Each finding in the ReXGrounding challenge test set that the CTLink ensemble missed (Dice score $<$ 0.1) was assigned automatically to one of four error types:
an empty prediction, partial overlap with the reference (Dice score above 0 but below 0.1), a prediction in the lobe of
the reference without overlap (a different lesion in the right place), or a prediction outside that lobe (wrong
location).

\subsection*{Statistical Analysis}
95\% CIs were estimated as percentile intervals from 2000 bootstrap replicates resampling scans. Models were compared
with two-sided paired bootstrap tests on the same resampled scans, and $P < .05$ indicated a statistically significant
difference.
Models were implemented in Python (version 3.11), PyTorch (version 2.10), and nnU-Net (version 2.8.1), and statistical
analyses were performed with NumPy (version 2.4).

\section*{Results}

\subsection*{Dataset Characteristics}
The v1 training set comprised 2989 scans from 2714 patients (mean age, 41.9 years $\pm$ 14.3 [SD]; 56\% male), the
v2 training set 1134 scans from 1078 patients (41.5 years $\pm$ 14.3; 57\% male), the validation set 200 scans from 190
patients (39.5 years $\pm$ 13.0; 56\% male), and the test set 300 scans from 281 patients (42.8 years $\pm$ 14.7;
58\% male). Across the splits, 55\%--60\% of scans were acquired with Philips, 32\%--36\% with Siemens, and 8\%--9\% with
Philips--Neusoft scanners. Most findings were focal: 6055 of 7652 (79\%) in the v1 training set, 1704 of 1895 (90\%) in the v2
training set, 329 of 381 (86\%) in the validation set, and 509 of 582 (87\%) in the test set (Table~\ref{tab:data}).

\subsection*{Ablation on the Validation Set}
Table~\ref{tab:ablation} shows the cumulative ablation. Most of the improvement over the published VoxTell model came
from the training data: fine-tuning on the v1 scans with corrected masks raised the hit rate from 0.533 to 0.761, and
adding the exhaustively annotated v2 scans raised it to 0.793 ($P = .03$) and the mean Dice score from 0.353 to 0.388
($P < .001$). A comparison of the original and corrected v1 masks is shown in Appendix~\ref{app:v1} and Table~\ref{tab:supp_v1}.
Post-processing did not change the hit rate ($P = .24$) but raised the instance F1 score from 0.373 to
0.456 ($P < .001$), because it removed spurious components (instance precision, 0.293 to 0.460). Overall, CTLink localized 82\% of validation findings,
compared with 53\% for VoxTell.

\subsection*{Challenge Test Set}
Table~\ref{tab:challenge} compares CTLink with the ReXGrounding challenge entries on all 300 test scans. A single
CTLink model matched the challenge winner in hit rate (0.766 vs 0.768; $P = .99$) and mean Dice score (0.364 vs 0.375;
$P = .14$) and had a higher instance F1 score (0.378 vs 0.341; $P = .005$), mainly because post-processing removed
small spurious components. The CTLink ensemble
had the highest hit rate (0.782) and instance F1 score (0.386) of all entries and the second-highest Dice
score (0.373). Compared with the winner, the difference in
instance F1 score was significant ($P = .001$), whereas the differences in hit rate ($P = .41$) and
Dice score ($P = .71$) were not. Table~\ref{tab:supp_test} gives precision and recall, and Table~\ref{tab:supp_agesex} an analysis by patient sex and age (Appendix~\ref{app:agesex}).

\begin{figure}[!htbp]\centering
\includegraphics[width=0.9\linewidth]{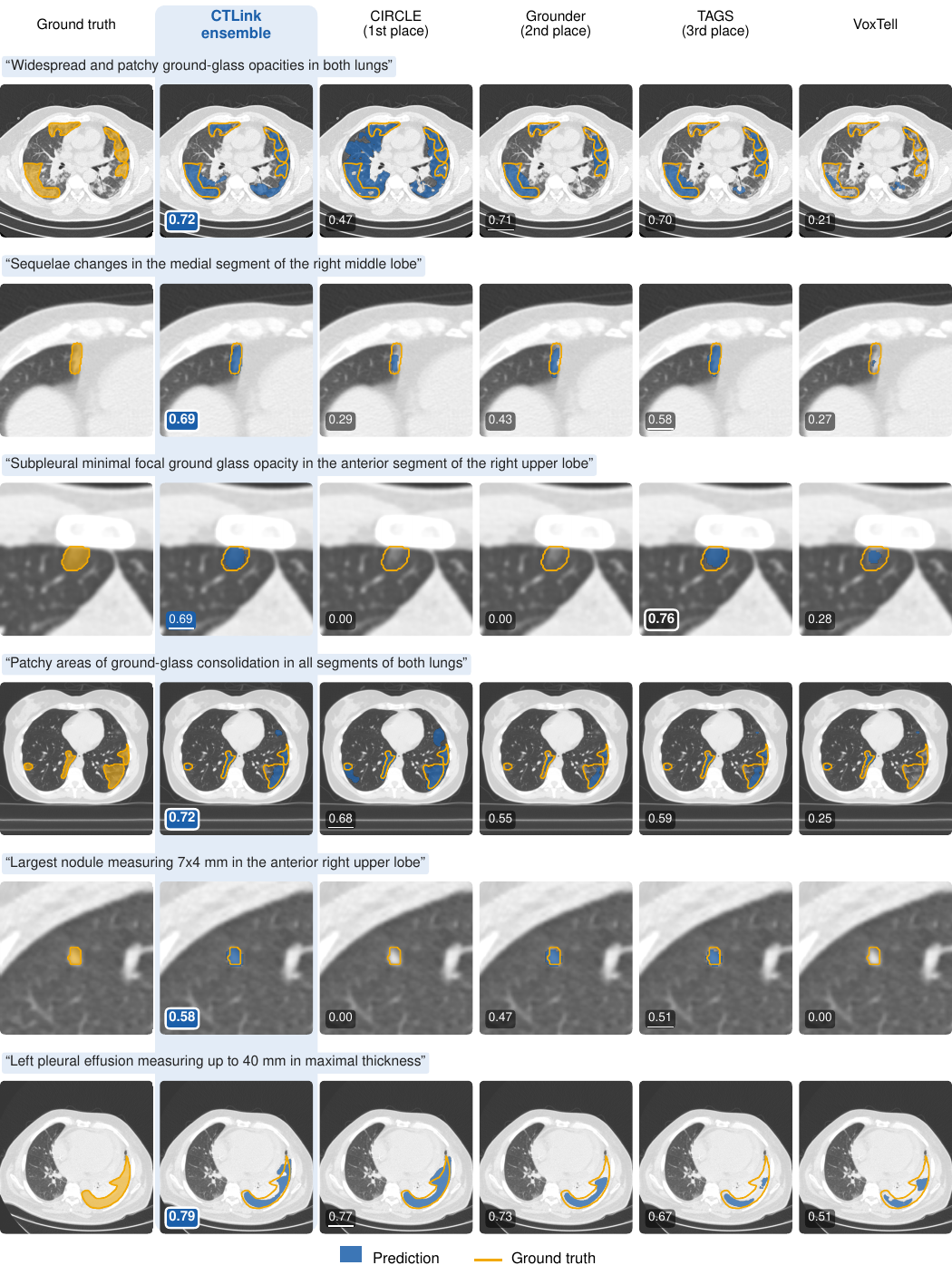}
\caption{Examples from the ReXGrounding challenge test set comparing the CTLink ensemble with the three top-ranked
challenge entries \cite{baharoon2026grounding} and VoxTell \cite{rokuss2026voxtell}. Numbers are Dice scores over the whole scan; the highest in each row is outlined and the second highest
underlined.}
\label{fig:qual}
\end{figure}

\begin{figure}[!htbp]\centering
\includegraphics[width=0.84\linewidth]{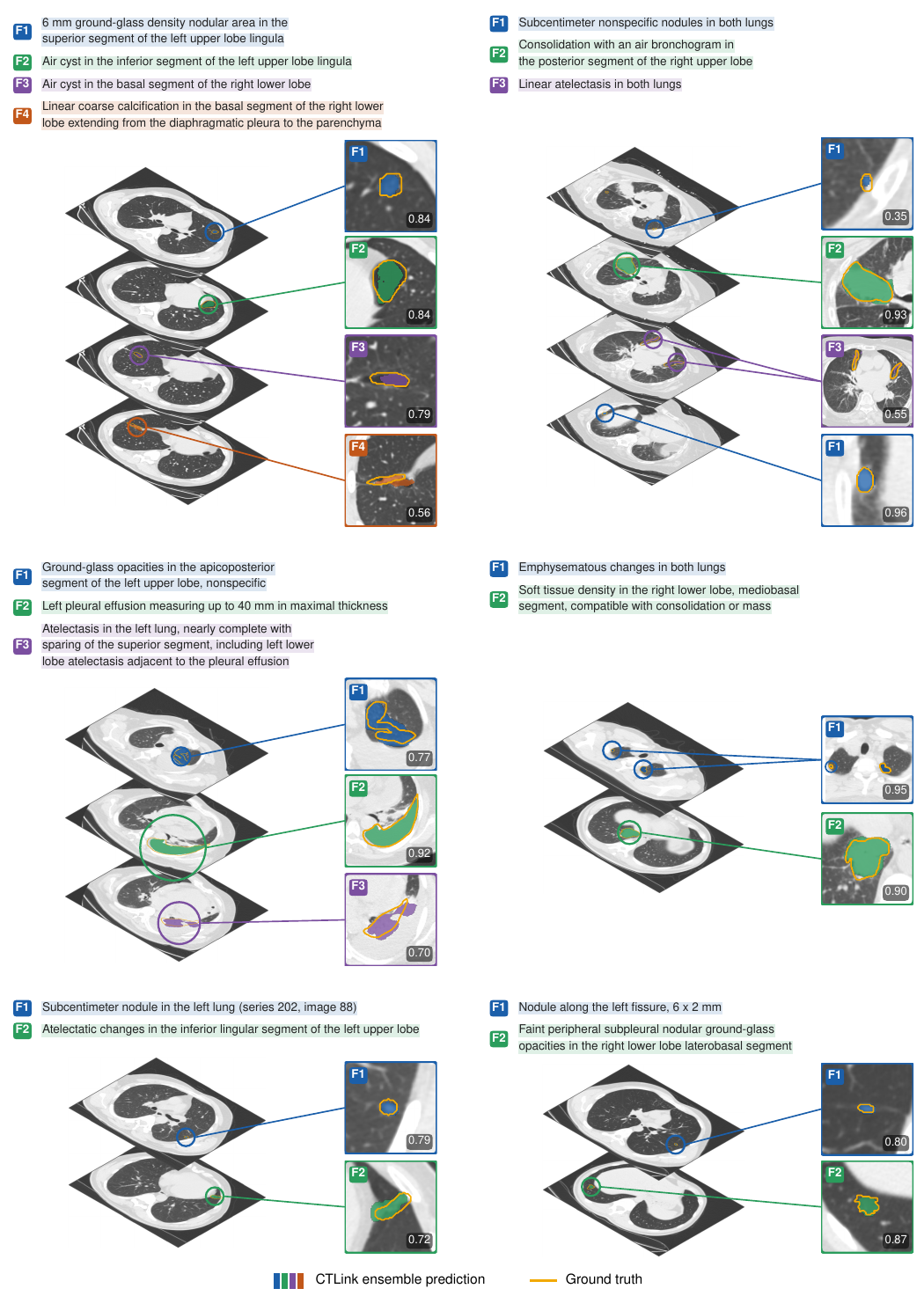}   %
\caption{Examples of successful CTLink ensemble segmentations on the ReXGrounding challenge test set. Numbers
are Dice scores on the shown section.}
\label{fig:multi}
\end{figure}

Figure~\ref{fig:qual} compares the CTLink ensemble with the three top-ranked challenge entries and VoxTell on six test
findings, and Figure~\ref{fig:multi} shows examples of successful CTLink ensemble segmentations.

\subsection*{Comparison with Radiologists}
On the 50 test scans with a second segmentation (96 findings;
Table~\ref{tab:interrater}), CTLink had a higher mean Dice
score than the second segmentation (0.374 vs 0.321; $P = .02$), with a similar hit rate (0.823 vs 0.760;
$P = .20$) and instance F1 score (0.399 vs 0.417; $P = .60$). The CTLink ensemble had a higher hit rate (0.854;
$P = .045$) and mean Dice score (0.381; $P = .01$) than the second segmentation and a similar instance F1 score
(0.412; $P = .88$). Of the 17 non-focal findings, the second segmentation localized 7 (41\%), CTLink 11 (65\%),
and the CTLink ensemble 12 (71\%).
With the second segmentation as the reference, CTLink reached a mean Dice score of 0.381 (CTLink
ensemble, 0.380), compared with 0.321 between the two segmentations ($P = .008$), and a similar hit rate (0.771 vs
0.760; $P = .88$). CTLink thus matched or exceeded inter-rater agreement.

\begin{table}[!htbp]
\footnotesize
\caption{\textbf{Comparison with a Second Radiologist Segmentation on 50 Test Scans.} Data are values (95\% CIs) for the 96 findings that the second segmentation also localized in the 50 ReXGrounding challenge test scans segmented twice in the inter-rater study of ReXGroundingCT \cite{baharoon2026rexgroundingct}. Reference masks are those of the challenge, and the second segmentation was drawn independently by other radiologists in the inter-rater study. Reference refers to the masks each segmentation was scored against. All masks were scored with the challenge evaluator. Mean Dice score per scan.}
\setlength{\tabcolsep}{3pt}
\begin{tabular*}{\linewidth}{@{\extracolsep{\fill}}llccc@{}}
\toprule
Reference & Segmentation & Instance F1 & Dice & Hit Rate \\
\midrule
Reference masks & Second segmentation & 0.417 (0.358, 0.478) & 0.321 (0.264, 0.379) & 0.760 (0.674, 0.839) \\
 & CTLink & 0.399 (0.343, 0.461) & 0.374 (0.316, 0.434) & 0.823 (0.750, 0.893) \\
 & CTLink ensemble & 0.412 (0.355, 0.473) & 0.381 (0.325, 0.440) & 0.854 (0.785, 0.919) \\
\midrule
Second segmentation & CTLink & 0.460 (0.394, 0.538) & 0.381 (0.327, 0.435) & 0.771 (0.697, 0.845) \\
 & CTLink ensemble & 0.458 (0.387, 0.534) & 0.380 (0.327, 0.434) & 0.771 (0.697, 0.845) \\
\bottomrule
\end{tabular*}
\label{tab:interrater}
\end{table}

\subsection*{Performance by Finding Category and Lesion Size}
\begin{figure}[!htbp]\centering
\includegraphics[width=0.95\linewidth]{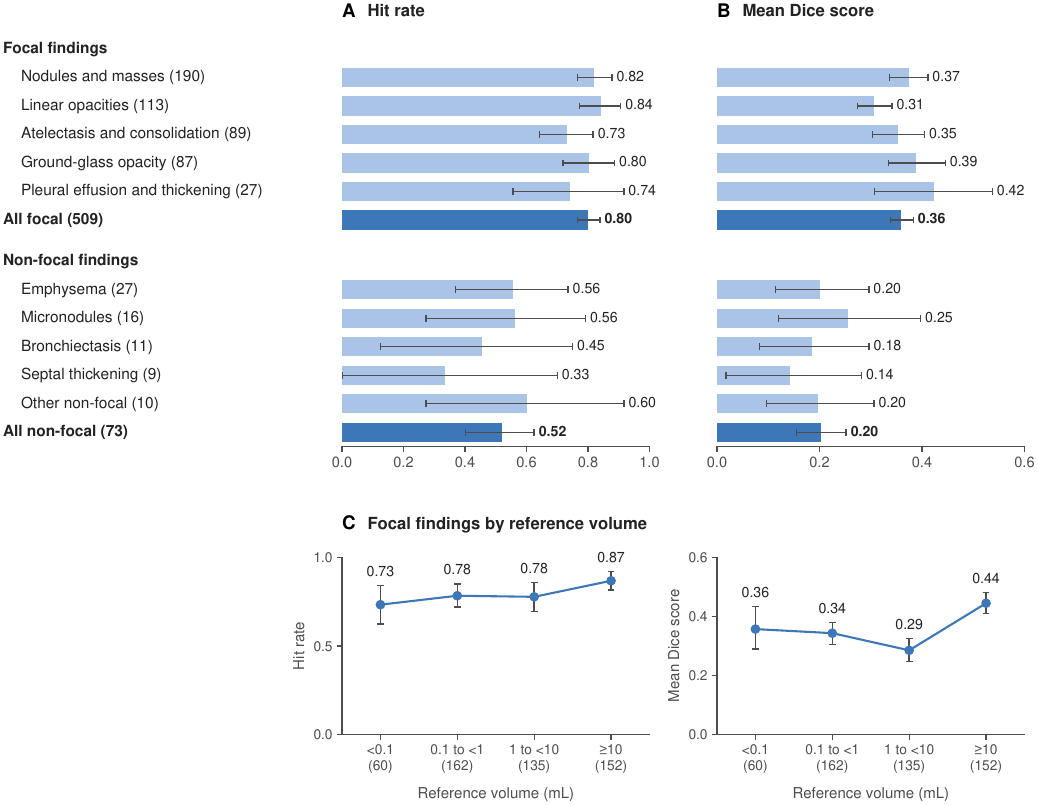}
\caption{Performance of CTLink by finding category and lesion size on the ReXGrounding challenge test set. (A) Hit rate and (B) mean Dice score of the findings in each focal and non-focal category, with the number of
findings in parentheses. (C) Hit rate and mean Dice score of focal findings by reference volume, with the number of findings in parentheses. ``Other non-focal'' includes bronchial wall thickening and uncategorized non-focal findings. ``All focal''
also includes one pneumothorax and two ``other'' focal findings. Error bars show 95\% CIs (2000 bootstrap replicates resampling scans).}
\label{fig:categories}
\end{figure}

On the test set, CTLink localized focal findings more often than non-focal findings (Figure~\ref{fig:categories}). The hit rate ranged from 0.73 for atelectasis and consolidation ($n$ = 89) to 0.84 for
linear opacities ($n$ = 113) across focal categories and from 0.33 for septal thickening ($n$ = 9) to 0.60 for other
non-focal findings ($n$ = 10) across non-focal categories. Among focal findings, the hit rate
was similar for all reference volumes below 10 mL and higher for findings of 10 mL or more (0.87; Figure~\ref{fig:categories}C), and only 16 of the 101
missed focal findings were smaller than 0.1 mL.

\subsection*{Robustness Analysis}
\begin{figure}[!htbp]\centering
\includegraphics[width=0.675\linewidth]{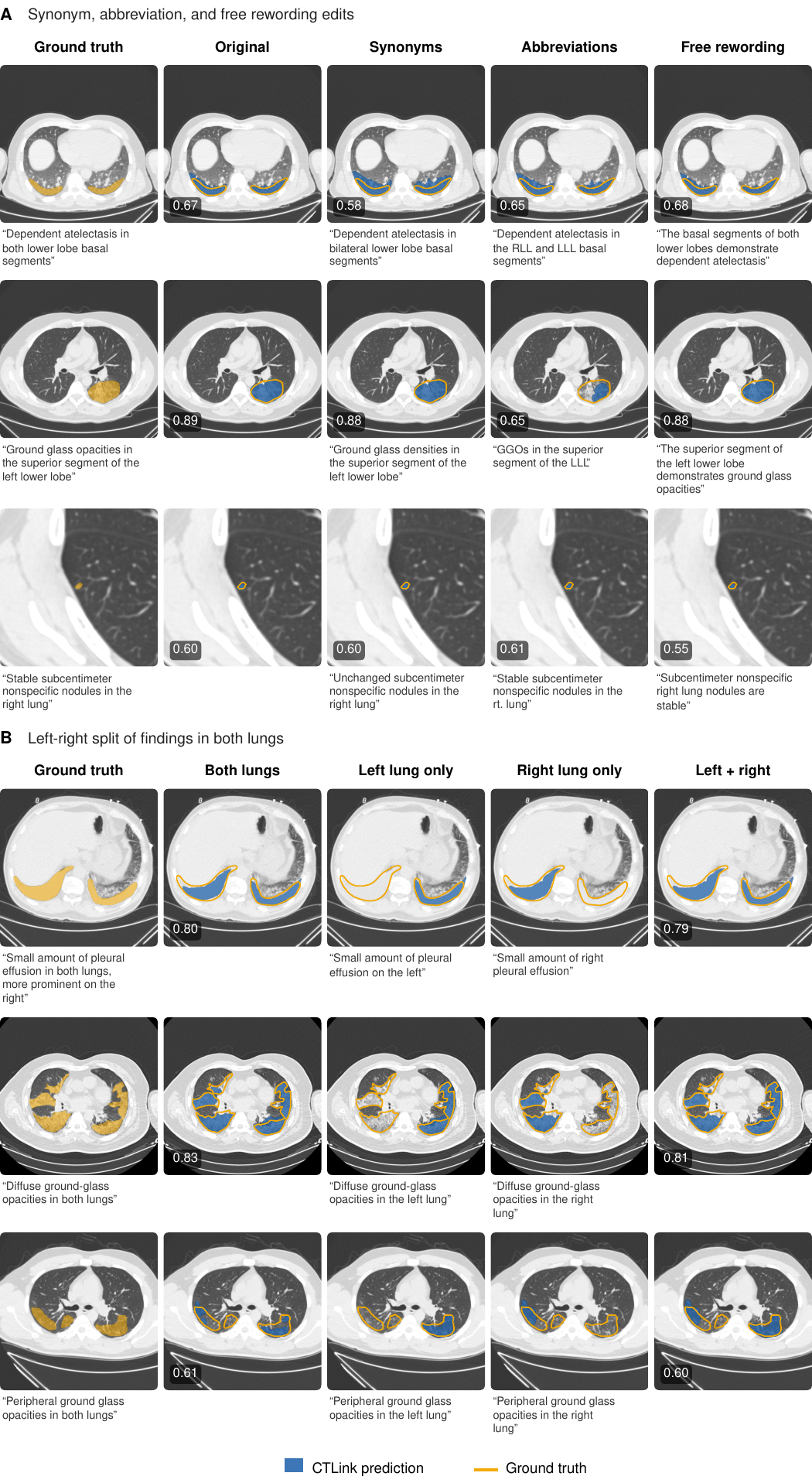}   %
\caption{Robustness analysis of CTLink without post-processing on the ReXGrounding challenge test set. (A) The same finding
requested with the original sentence (under the ground truth), with synonyms, with abbreviations, and in free wording. (B)
Findings in both lungs requested at once and for each lung separately. Images are in radiological orientation, with the
patient's right on the left of the image. Numbers are Dice scores over the whole scan.}
\label{fig:wording}
\end{figure}

Across the four edit types, CTLink was robust to the wording of the
finding (Figure~\ref{fig:wording}; Table~\ref{tab:supp_wording}). With synonyms or free rewording, its masks agreed
closely with the original masks (median Dice score, 0.95 and 0.91), and the hit rate did not change
significantly ($P = .71$ and $P = .92$). Only abbreviations lowered the hit rate, by 0.043 ($P < .001$). For the 290
findings in both lungs, requesting each lung separately and taking the union did not change the hit rate (0.762 vs 0.766; $P = .89$), and
even without post-processing, 93\% of the voxels predicted for a single lung lay in the requested lung.
For the 31 findings in two or more lobes of one lung, requesting each lobe separately gave a hit rate of 0.806, compared with 0.710 for the
original sentence ($P = .07$), and without post-processing, 78\% of the voxels predicted for a single lobe lay in the requested lobe.

\subsection*{Error Analysis}
\begin{figure}[!htbp]\centering
\includegraphics[width=0.9\linewidth]{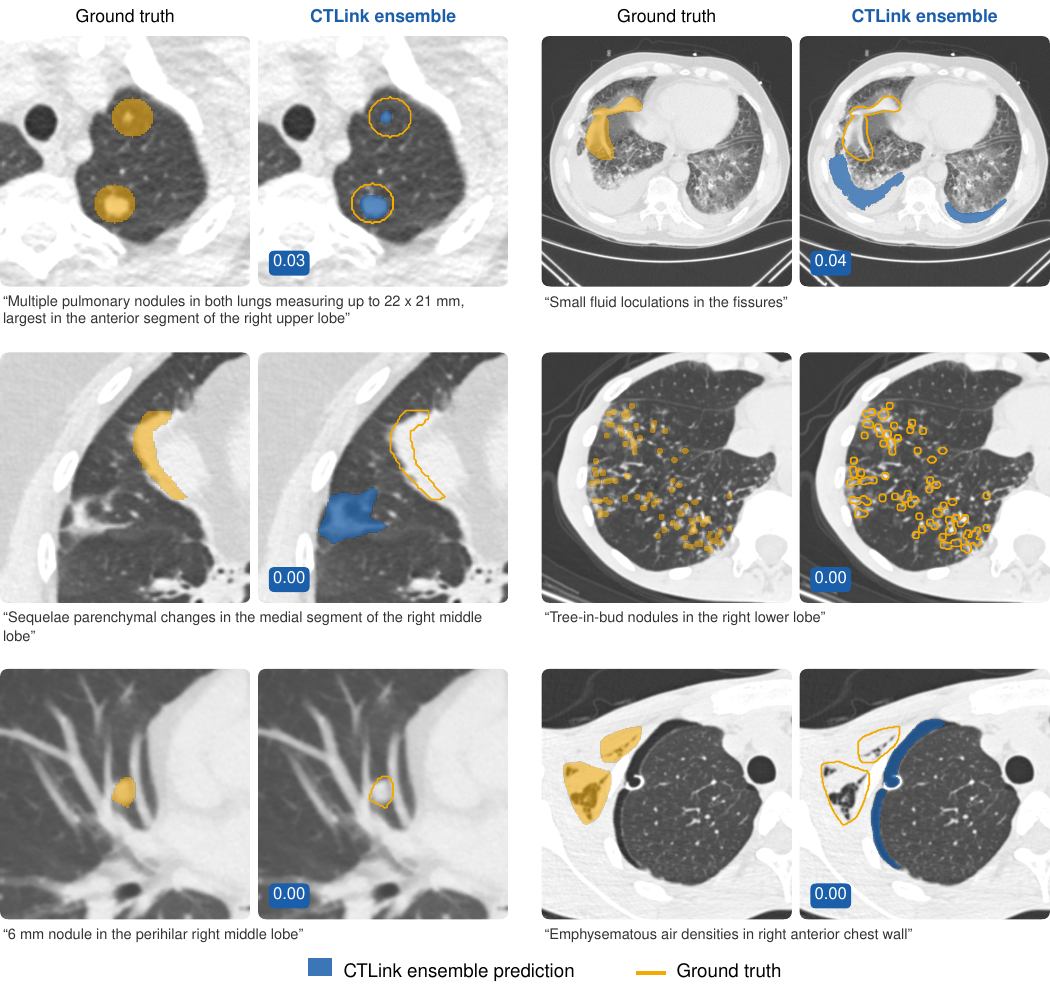}
\caption{Examples of findings that the CTLink ensemble missed on the ReXGrounding challenge test set. Each example shows
the ground truth and the CTLink ensemble prediction on the same section. Numbers are Dice scores over
the whole scan. Top left, on this section the ensemble segmented the nodules themselves, whereas the reference masks
extend beyond them. Top right, it segmented the pleural fluid at the lung bases instead of the fluid in the fissures.
Bottom right, it segmented the pneumothorax along the right lung instead of the air in the chest wall.}
\label{fig:failures}
\end{figure}

Of the 127 test findings that the CTLink ensemble missed (examples in Figure~\ref{fig:failures}), 3 (2\%) had an empty prediction, 75 (59\%) overlapped the
reference partially, 22 (17\%) had a prediction in the lobe of the reference without overlap, and 27 (21\%) had a
prediction in another location. Partial overlap was the most common error type for both focal (58 of 97, 60\%) and
non-focal findings (17 of 30, 57\%). Some misses reflected the reference masks or a related abnormality rather than a failure to localize (Figure~\ref{fig:failures}). For ``Multiple pulmonary nodules in both lungs,'' the reference masks extended beyond the nodules that the CTLink ensemble segmented. For ``Small fluid loculations in the fissures,'' it segmented the pleural fluid at the lung bases, and for ``Emphysematous air densities in right anterior chest wall,'' it segmented the pneumothorax along the right lung instead of the air in the chest wall.

\section*{Discussion}
We developed CTLink, a text-prompted segmentation model for linking radiology report findings to 3D chest CT scans. To train it, we created ReXGroundingCTv2, which adds 1134 exhaustively annotated scans to ReXGroundingCT. CTLink localized 82\% of findings on
the ReXGroundingCT validation set, matched or exceeded radiologist inter-rater agreement on 50 test scans, and an ensemble of CTLink models had the highest hit rate and instance F1 score of all entries of the ReXGrounding
challenge. At this level, CTLink could mark most reported findings on the scan for referring clinicians,
and even most of the findings that the CTLink ensemble missed (75 of 127) overlapped the reference partially. The ablation showed that training data drove most of these gains: fine-tuning VoxTell on the
v1 training set, which had been only a small part of its pretraining, gave by far the largest improvement, and adding
the exhaustively annotated v2 training set the largest further gain in Dice score.

Our study had several limitations. First, non-focal findings, such as emphysema and bronchiectasis, remained the hardest part
of the task, partly because the extent of a diffuse finding is hard to outline consistently. In the inter-rater study
of ReXGroundingCT, the second radiologist segmentation reached a hit rate of only 0.41 for non-focal findings against the
reference masks, compared with 0.84 for focal findings \cite{baharoon2026rexgroundingct}, and reader agreement on diffuse lung disease at CT is moderate
even for visual assessment \cite{watadani2013interobserver,delaney2024meta}. Second, all scans came from CT-RATE, a single-center collection of
noncontrast chest CT scans, so performance on contrast-enhanced scans, other scanners, and other populations is
unknown. Third, each reference mask was drawn by one annotator, although readers disagree even on which lung nodules
are present \cite{armato2011lung}. \new{Fourth}, the comparison
with radiologists covered only 50 scans (96 findings). \new{Finally, CTLink covers only findings of the lungs, airways, and pleura.}

In conclusion, CTLink localized the findings described in chest CT reports at the level of agreement between
radiologists, showing that report sentences can be linked automatically to the abnormalities they describe. Such
links could let referring clinicians see each reported finding on the scan, make it easier to compare findings with
prior examinations, and let readers verify AI-generated reports. Our results also showed that
exhaustive and accurate annotation of the training data mattered more than changes to the training procedure, which
should guide the design of future datasets.
Future work should
extend the approach to other body regions and imaging modalities, and establish consistent
annotation of diffuse findings. We release CTLink and ReXGroundingCTv2 to support this work.

\section*{Acknowledgments}
We thank Bayan Altalaq, Sakinah Alabdrabalnabi, Seham Almalki, and Anthony Tran for contributing annotations to ReXGroundingCTv2.

\section*{Funding}
This study did not receive any specific grant from funding agencies in the public, commercial, or not-for-profit
sectors.

\section*{Data and Code Availability}
The CT scans and reports are available from CT-RATE \cite{hamamci2026generalist}. The ReXGroundingCT annotations,
including the challenge validation and test reference masks, and the challenge evaluation code are available at
\url{https://huggingface.co/datasets/rajpurkarlab/ReXGroundingCT} under a CC BY-NC-SA 4.0 license. ReXGroundingCTv2,
which adds the exhaustive annotations of the 1134 training scans introduced in this study and
includes the corrected v1 training masks, is released at the same address under the same license. The CTLink code is available at \url{https://github.com/rajpurkarlab/CTLink}, and the model weights at \url{https://huggingface.co/rajpurkarlab/CTLink}.

\section*{Declaration of Generative AI and AI-Assisted Technologies in the Manuscript Preparation Process}
During the preparation of this work the authors used Claude (Opus 5.5 and Fable 5.2, Anthropic) to help draft and edit
the text and write analysis and figure code. The authors reviewed and edited the content as needed and take full responsibility for the content of the published
article.

\bibliographystyle{unsrtnat}
\bibliography{refs}

\clearpage
\section*{Supplemental Material}
\setcounter{table}{0}\renewcommand{\thetable}{S\arabic{table}}\setcounter{figure}{0}\renewcommand{\thefigure}{S\arabic{figure}}

\appsection{Annotators}{app:annot}
The v1 training masks were drawn by professional annotators or trained medical students and reviewed by board-certified
radiologists \cite{baharoon2026rexgroundingct}. The final annotations of the 1134 ReXGroundingCTv2 training scans were made by board-certified radiologists for 887
scans (78.2\%), by radiology residents for 226 scans (19.9\%), and by medical students for 21 scans (1.9\%).
All reviewers were board-certified radiologists, who could accept, edit, or
reject an annotation. All 21 scans finalized by medical students and 96 of the 226 scans finalized by radiology residents
were reviewed, so the annotations of 1004 scans (88.5\%) were drawn or reviewed by a board-certified radiologist.

\appsection{Instance F1 Score of the Second Segmentation}{app:f1r2}
Table~\ref{tab:interrater} gives an instance F1 score of 0.417 for the second segmentation, whereas the inter-rater
study of ReXGroundingCT reported 0.47 for the same masks \cite{baharoon2026rexgroundingct}. The difference comes from how
instances are defined. The inter-rater study of ReXGroundingCT used the instances as the radiologists drew them. The
challenge evaluator, which we used for every model and for the second segmentation, instead counts each connected
component of a mask as one instance, so a lesion drawn with disconnected parts counts as several instances. Scoring all masks with the evaluator keeps CTLink and the radiologists comparable. Hit
rate and Dice score do not depend on instances and are the same under both definitions.

\appsection{Robustness Analysis}{app:rob}
The rewritten sentences were generated with GPT-6.1-sol (version 2026-09-29; OpenAI) and checked with GPT-6-astra (version 2026-09-03; OpenAI), with the prompts in Figures~\ref{fig:prompt_gen} to~\ref{fig:prompt_lobe}. For each of the 582 test findings, the model wrote (1) a version in which terms were replaced only with synonyms that radiologists commonly write in their place with exactly the same meaning (for example, ``Dependent atelectasis in both lower lobe basal segments'' became ``Dependent atelectasis in bilateral lower lobe basal segments''), (2) a version with standard radiology abbreviations (such as RUL, GGO, and bilat.), and (3) a version freely reworded as another radiologist would write it. (4) It also judged whether the finding involved both lungs (307
findings) and, if so, wrote a left-lung and a right-lung sentence that kept the finding type, the descriptors that apply
to every lesion, and the locations on that side (for example, ``Scarring in the apical segments of the upper lobes of both lungs'' became ``Scarring in
the apical segment of the left upper lobe'' and ``Scarring in the apical segment of the right upper lobe''). (5) In the same way, it decided which lobes the finding involved, as stated or clearly implied by the sentence, and, for a finding in two or more lobes, wrote one sentence per lobe (34 findings involved two or more lobes of one lung).

A sentence was analyzed only if the check accepted it and it differed from the original. This left 297 of the 582 synonym sentences (the other 285 had no such synonym), 540 of the 582 abbreviated sentences, all 582 freely reworded sentences, and the left-lung and right-lung sentences of 290 of the 307 bilateral findings. For the lobe split, all lobe sentences were accepted for 31 of the 34 findings.

Each sentence was segmented by CTLink with the same post-processing. The lobe restriction used the lobes named in the
sentence given, so it did not apply to abbreviated lobe names. CTLink without post-processing shows the effect of the
network alone. Agreement was the Dice score between
the masks predicted for the original and the reworded sentence (1 when both were empty). For bilateral findings, the
left-lung and right-lung predictions were combined into one mask by taking their union and scored with the challenge evaluator against the
reference mask of the finding. The side of each predicted voxel was that of the nearer lung in the TotalSegmentator
lobe masks. The predictions for a finding's lobe sentences were combined in the same way, and the lobe of each predicted voxel was that of the nearest TotalSegmentator lobe. 95\% CIs and $P$ values are from paired bootstrap tests resampling scans (2000 replicates).

\subsubsection*{Prompts}
The prompts are shown in Figures~\ref{fig:prompt_gen} to~\ref{fig:prompt_lobe}, with the inserted text in braces.

\begin{figure}[H]\centering
\begin{promptfig}{Synonyms, Abbreviations, and Free Rewording}
You are a thoracic radiologist. Below is one finding sentence from a chest CT report. Write variants of it.\par\medskip
Rules for every variant: keep the meaning exactly - the same finding type, the same location (lung, lobe, segment), the same
size, count, number (singular/plural) and descriptors. Do not add or remove information. Write one sentence in the style of a
CT report.\par\medskip
Return a JSON object with these keys:
\begin{itemize}[label=-, leftmargin=1.2em, nosep]
\item ``syn'': Rewrite the sentence replacing medical terms with synonyms that radiologists typically use in chest CT reports. Replace a term
only if radiologists commonly write the synonym in place of it with exactly the same meaning; do not make any term more or less
specific (e.g. keep `nodule' as a nodule, do not turn `scarring' into `fibrosis' or `multiple' into `numerous') and do not add
or remove descriptors. Keep every term that has no such commonly used synonym unchanged. Keep the sentence structure. The
result must read like a sentence a radiologist would write in a chest CT report. If no term can be replaced this way, return
the original sentence unchanged.
\item ``abbr'': the same sentence written with standard radiology abbreviations where natural (e.g. RUL, RML, RLL, LUL, LLL,
GGO, bilat.).
\item ``reword'': the same finding written freely, as a different radiologist would write it (new word order and phrasing).
\end{itemize}\medskip
Sentence: ``\promptvar{sentence}''\\
Return only the JSON object.
\end{promptfig}
\caption{Instructions used to write the synonym, abbreviated, and freely reworded sentences.}
\label{fig:prompt_gen}
\end{figure}

\begin{figure}[H]\centering
\begin{promptfig}{Left-Right Split}
You are a thoracic radiologist. Original chest CT finding sentence: ``\promptvar{sentence}''\par\medskip
Task: Rewrite this bilateral finding as the \promptvar{side}-lung part only.\par\smallskip
Rules:
\begin{itemize}[label=-, leftmargin=1.2em, nosep]
\item keep the finding type and descriptors that apply to every lesion (e.g. `subcentimeter', `indeterminate',
`calcified' when all are calcified);
\item keep only \promptvar{side}-sided locations stated or implied for that side (e.g. `both lower lobes' $\rightarrow$
`\promptvar{side} lower lobe', `apices' $\rightarrow$ `\promptvar{side} lung apex');
\item DROP counts and quantifiers that describe both lungs together (`several', `multiple', `a few', `one or two', `two to
three'), size maxima and `largest \ldots' clauses unless the original places them on the \promptvar{side} side, and partial
qualifiers (`some calcified', `asymmetric') that cannot be assigned to one side;
\item turn plural cross-references to the other side (e.g. `pleural effusions') into the \promptvar{side}-sided singular.
\end{itemize}\smallskip
Keep the noun plural if the original noun is plural. Mention nothing about the other lung.\par\medskip
Return a JSON object \{``text'': ``<the new sentence>''\}.
\end{promptfig}
\caption{Instructions used to write the left-lung and right-lung sentences of the findings that the model judged to involve both lungs (stated or implied, for example ``bilateral'', ``both lungs'', or lobes on both sides).}
\label{fig:prompt_side_task}
\end{figure}

\begin{figure}[H]\centering
\begin{promptfig}{Lobe Split}
You are a thoracic radiologist. Below is one finding sentence from a chest CT report.\par\medskip
Decide which lung lobes the finding involves, as stated or clearly implied by the sentence (the lingula belongs to the left
upper lobe; ``both lower lobes'' means the right lower lobe and the left lower lobe). Do not assign lobes from general locations
such as ``both lungs'', ``peripheral'', ``subpleural'' or ``basal'' unless the sentence names or clearly implies the
lobes.\par\medskip
If the finding involves two or more lobes, write, for each involved lobe, one sentence that describes only the part of the
finding in that lobe: keep the finding type and the descriptors that apply to every lesion, keep the locations within that lobe
(for example segments) that the original states, name the lobe explicitly, and drop counts, size maxima and comparisons that
describe several lobes together. Do not add anything. Use null for lobes that the finding does not involve.\\
If the finding involves fewer than two lobes, or its lobes are not stated or clearly implied, return null for every
lobe.\par\medskip
Sentence: ``\promptvar{sentence}''\\
Return only a JSON object with the keys ``RUL'', ``RML'', ``RLL'', ``LUL'', ``LLL'' (right upper, right middle, right lower,
left upper, left lower lobe).
\end{promptfig}
\caption{Instructions used to decide which lobes a finding involves and to write the lobe sentences.}
\label{fig:prompt_lobe}
\end{figure}

\appsection{Original versus Corrected v1 Masks}{app:v1}
To measure the effect of the corrected v1 masks, VoxTell was fine-tuned on the v1 scans twice, once with the original
and once with the corrected v1 masks, with the same recipe and random seed. Both models were evaluated at their final checkpoint on the 200 validation scans without post-processing
(Table~\ref{tab:supp_v1}).

\appsection{Performance by Patient Sex and Age}{app:agesex}
Patient sex and age were taken from the CT-RATE metadata (sex per patient, age at the scan). Both were available for all
300 test scans. On the 300 ReXGrounding
challenge test scans, CTLink had a lower hit rate (0.740 vs 0.802) and instance F1 score (0.367 vs 0.395) in men than in women, but the differences
were not statistically significant ($P = .09$ and $P = .30$), and the mean Dice score was the same in both (0.364; $P = .99$;
Table~\ref{tab:supp_agesex}).
In patients 60 years or older, CTLink had a lower hit rate (0.748 vs 0.799) and mean Dice score (0.355 vs 0.396) than in those
younger than 40 years, but the differences were also not statistically significant ($P = .30$ and $P = .22$). Its instance F1 score was
also lower in patients 60 years or older (0.337 vs 0.413; $P = .03$), who had more findings per scan (2.6 vs 1.8).

\clearpage
\subsection*{Supplemental Tables}

\begin{table}[h]
\footnotesize
\caption{\textbf{Original versus Corrected v1 Masks on the Validation Set.} Data are values for the ReXGrounding challenge validation set. Both models were fine-tuned from VoxTell on the v1 scans only, with the same recipe and random seed, and differ only in the v1 masks: original (2992 scans, 7687 findings) or corrected (2989 scans, 7652 findings). Dice is the mean per-scan Dice score.}
\begin{tabular*}{\linewidth}{@{\extracolsep{\fill}}lccc@{}}
\toprule
v1 Masks & Instance F1 & Dice & Hit Rate \\
\midrule
Original & 0.281 & 0.346 & 0.745 \\
Corrected & 0.298 & 0.353 & 0.761 \\
\bottomrule
\end{tabular*}
\label{tab:supp_v1}
\end{table}

\begin{table}[h]
\small
\caption{\textbf{Precision and Recall of CTLink and the Challenge Entries on the Test Set.} Data are values for the ReXGrounding challenge test set. Instance precision and recall match connected components of the prediction one to one with the annotated reference instances (match at Dice score $\geq$ 0.2; predicted components smaller than 10 voxels ignored). Hit recall is the proportion of findings with a Dice score of at least 0.1 (the hit rate), and hit precision is the proportion of findings with a nonempty prediction that were hits.}
\setlength{\tabcolsep}{3pt}
\begin{tabularx}{\linewidth}{@{}>{\raggedright\arraybackslash}p{0.42\linewidth}YYYY@{}}
\toprule
& \multicolumn{2}{c}{Instance} & \multicolumn{2}{c}{Hit} \\ \cmidrule(lr){2-3}\cmidrule(l){4-5}
Entry & Precision & Recall & Precision & Recall \\
\midrule
CTLink without post-processing & 0.178 & \textbf{0.514} & 0.787 & 0.773 \\
CTLink & \textbf{0.336} & 0.432 & 0.789 & 0.766 \\
CTLink ensemble & 0.335 & 0.455 & \textbf{0.793} & \textbf{0.782} \\
\midrule
Team 1 (winner) & 0.272 & 0.457 & 0.786 & 0.768 \\
Team 2 & 0.280 & 0.443 & 0.766 & 0.763 \\
Team 3 & 0.254 & 0.487 & 0.770 & 0.765 \\
Team 4 & 0.242 & 0.438 & 0.779 & 0.777 \\
Team 5 & 0.117 & 0.497 & 0.746 & 0.746 \\
Team 6 & 0.162 & 0.492 & 0.769 & 0.765 \\
Team 7 & 0.183 & 0.456 & 0.764 & 0.753 \\
Team 8 & 0.169 & 0.456 & 0.753 & 0.742 \\
Team 9 & 0.169 & 0.466 & 0.724 & 0.716 \\
Team 10 & 0.231 & 0.421 & 0.702 & 0.689 \\
\bottomrule
\end{tabularx}
\label{tab:supp_test}
\end{table}

\begin{table}[h]
\footnotesize
\caption{\textbf{Performance by Patient Sex and Age on the Test Set.} Data are values for the ReXGrounding challenge test set. Mean Dice score per scan.}
\setlength{\tabcolsep}{3pt}
\begin{tabular*}{\linewidth}{@{\extracolsep{\fill}}lcccccccc@{}}
\toprule
 & & & \multicolumn{3}{c}{CTLink} & \multicolumn{3}{c}{CTLink Ensemble} \\
\cmidrule(lr){4-6}\cmidrule(l){7-9}
Patients & Scans & Findings & Instance F1 & Dice & Hit Rate & Instance F1 & Dice & Hit Rate \\
\midrule
\multicolumn{9}{@{}l}{\textit{Sex}} \\
\quad Male & 173 & 339 & 0.367 & 0.364 & 0.740 & 0.372 & 0.371 & 0.767 \\
\quad Female & 127 & 243 & 0.395 & 0.364 & 0.802 & 0.408 & 0.374 & 0.802 \\
\addlinespace
\multicolumn{9}{@{}l}{\textit{Age (years)}} \\
\quad $<$40 & 142 & 249 & 0.413 & 0.396 & 0.799 & 0.419 & 0.402 & 0.815 \\
\quad 40--59 & 113 & 218 & 0.365 & 0.328 & 0.739 & 0.373 & 0.341 & 0.766 \\
\quad $\geq$60 & 45 & 115 & 0.337 & 0.355 & 0.748 & 0.351 & 0.359 & 0.739 \\
\bottomrule
\end{tabular*}
\label{tab:supp_agesex}
\end{table}

\begin{table}[h]
\footnotesize
\caption{\textbf{Robustness Analysis: Effect of Phrasing.} Data are for the ReXGrounding challenge test set. Each edit is compared with the original ReXGroundingCT sentence on the findings whose sentence the edit changed. The left-right split covers the 290 findings in both lungs, with the left-lung and right-lung predictions combined. The lobe split covers the 31 findings in two or more lobes of one lung, with the per-lobe predictions combined. Agreement is the median Dice score between the masks predicted for the original and the edited sentence. Mean Dice score per scan.}
\begin{tabular*}{\linewidth}{@{\extracolsep{\fill}}lcccccccc@{}}
\toprule
 & & \multicolumn{2}{c}{Hit Rate} & \multicolumn{2}{c}{Dice} & \multicolumn{2}{c}{Instance F1} & \\
\cmidrule(lr){3-4}\cmidrule(lr){5-6}\cmidrule(lr){7-8}
Edit & Findings & Original & Edited & Original & Edited & Original & Edited & Agreement \\
\midrule
\multicolumn{9}{@{}l}{\textit{CTLink}} \\
\quad Synonyms & 297 & 0.771 & 0.768 & 0.370 & 0.366 & 0.385 & 0.382 & 0.95 \\
\quad Abbreviations & 540 & 0.774 & 0.731 & 0.363 & 0.349 & 0.380 & 0.368 & 0.88 \\
\quad Free rewording & 582 & 0.766 & 0.765 & 0.364 & 0.363 & 0.378 & 0.375 & 0.91 \\
\quad Left-right split & 290 & 0.766 & 0.762 & 0.332 & 0.323 & 0.347 & 0.326 & \textemdash \\
\quad Lobe split & 31 & 0.710 & 0.806 & 0.352 & 0.383 & 0.246 & 0.246 & \textemdash \\
\addlinespace
\multicolumn{9}{@{}l}{\textit{CTLink without post-processing}} \\
\quad Synonyms & 297 & 0.778 & 0.774 & 0.376 & 0.371 & 0.273 & 0.273 & 0.94 \\
\quad Abbreviations & 540 & 0.781 & 0.739 & 0.368 & 0.354 & 0.268 & 0.250 & 0.90 \\
\quad Free rewording & 582 & 0.773 & 0.768 & 0.369 & 0.367 & 0.264 & 0.261 & 0.91 \\
\quad Left-right split & 290 & 0.772 & 0.776 & 0.337 & 0.333 & 0.239 & 0.226 & \textemdash \\
\quad Lobe split & 31 & 0.710 & 0.871 & 0.364 & 0.386 & 0.150 & 0.171 & \textemdash \\
\bottomrule
\end{tabular*}
\label{tab:supp_wording}
\end{table}

\end{document}